\documentclass[10pt, a4paper]{article}

\usepackage{lrec-coling2024} 
\usepackage{multicol}
\usepackage{caption}
\usepackage{subcaption}

\title{Pre-Trained Language Models Represent \\ Some Geographic Populations Better Than Others}

\name{Jonathan Dunn$^1$, Benjamin Adams$^2$, and Harish Tayyar Madabushi$^3$} 

\address{$^1$University of Illinois Urbana-Champaign, Department of Linguistics \\
         $^2$University of Canterbury, Department of Computer Science and Software Engineering\\
         $^3$University of Bath, Department of Computer Science\\
         jedunn@illinois.edu, benjamin.adams@canterbury.ac.nz, htm43@bath.ac.uk\\}

\abstract{
This paper measures the skew in how well two families of LLMs represent diverse geographic populations. A spatial probing task is used with geo-referenced corpora to measure the degree to which pre-trained language models from the \textsc{opt} and \textsc{bloom} series represent diverse populations around the world. Results show that these models perform much better for some populations than others. In particular, populations across the US and the UK are represented quite well while those in South and Southeast Asia are poorly represented. Analysis shows that both families of models largely share the same skew across populations. At the same time, this skew cannot be fully explained by sociolinguistic factors, economic factors, or geographic factors. The basic conclusion from this analysis is that pre-trained models do not equally represent the world's population: there is a strong skew towards specific geographic populations. This finding challenges the idea that a single model can be used for all populations.
 \\ \newline \Keywords{spatial probing, population skew, geographic corpora, under-represented populations} }

\begin{document}

\maketitleabstract

\section{Introduction}

Large language models (LLMs) aim to provide a single set of representations that captures both linguistic knowledge and world knowledge across a diverse range of languages. Previous work has focused on developing probing tasks which measure the degree to which such models capture (i) purely linguistic knowledge (e.g., \citealt{10.3389/frai.2023.1225791}), (ii) reasoning abilities and world knowledge (e.g., \citealt{li-etal-2023-counterfactual}), and (iii) the ability to perform tasks across languages (e.g., \citealt{pires-etal-2019-multilingual}). A remaining question is whether such models work equally well across diverse populations using the same language. The study described in this paper first uses geography to demarcate different populations around the world and then uses comparable geo-referenced corpora to measure how well two families of LLMs perform across these different populations. Using a single \textit{lingua franca}, English, controls for potential confounds caused by observing different languages across different populations.

We focus on the following questions: 
\begin{itemize}
    \item What populations do pre-trained LLMs best describe, using perplexity scores on comparable corpora to measure goodness-of-fit?
    \item What are the geographic and social factors which best predict which populations are better represented by the models?
\end{itemize} 
We conduct this spatial probing task across models with an increasing number of parameters from two open-source series of LLMs: the BigScience \textsc{bloom} series (with 560m, 1.7 billion, and 3 billion parameters: \citealt{workshop2023bloom}) and the Facebook \textsc{opt} series (with 350 million, 1.7 billion, and 3 billion parameters: \citealt{zhang2022opt}). This approach allows us to determine whether larger models become increasingly more representative of the world's population and whether one family is itself more representative. If these are adequate general-purpose models of language then they would work equally well across all populations that use the same language (English) in the same register (social media). To the degree that a model deviates from this equal distribution of performance, it becomes a model of one population's language use rather than a model of language more broadly.

\begin{figure*}[t]
\centering
\includegraphics[width=.9\textwidth]{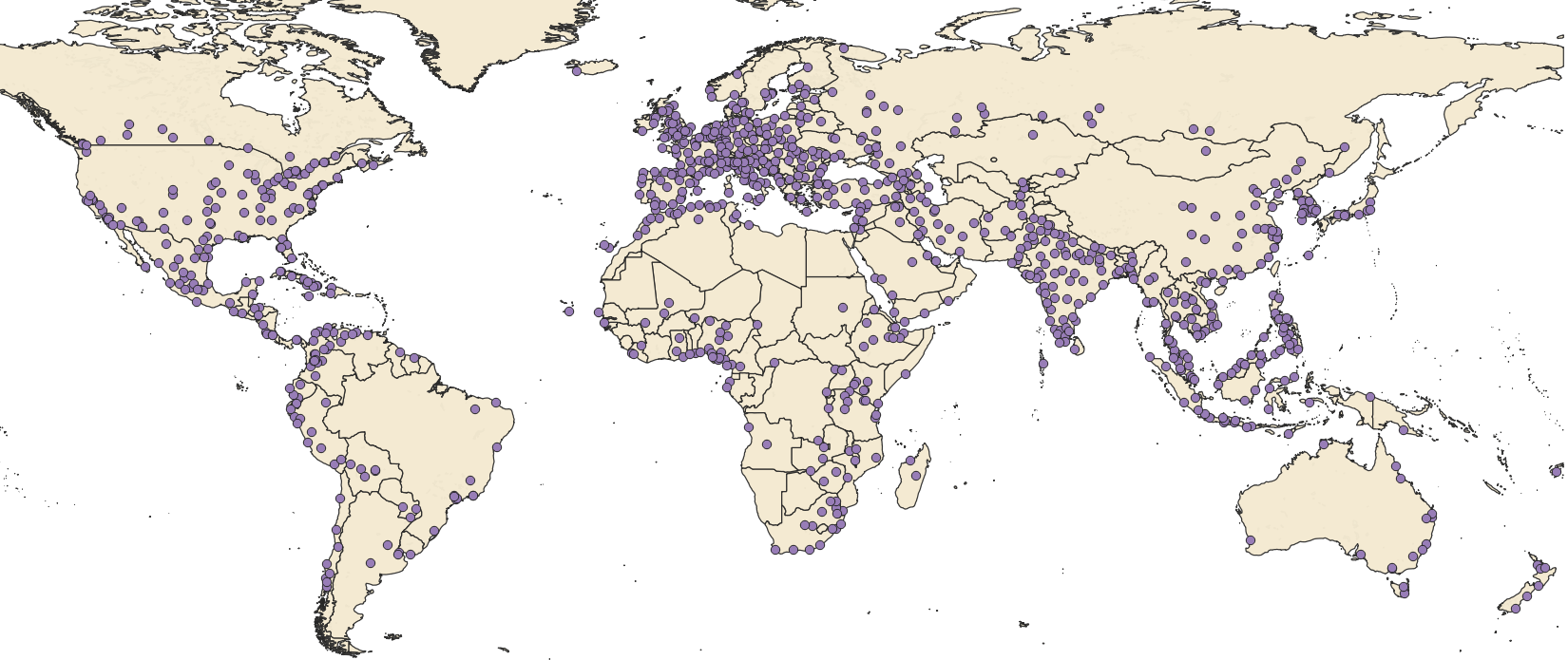}
\caption{Map of Local Populations. Each point represents a data collection location.}
\label{fig:map}
\end{figure*}

We first collect a corpus of social media that is balanced for key lexical items while representing the written production of 927 local populations. Using this geo-referenced corpus as a test set, the probing task uses the \textsc{bloom} and \textsc{opt} models to measure the perplexity of each local sub-corpus in order to determine whether the model finds comparable corpora around the world to be equally probable. This data set contains over 86,000 sub-corpora representing 130 countries. While we do not know the geographic sources of the training data for these models, this probing task allows us to reconstruct the likely distribution.

We then undertake a detailed analysis of the results of the probing task in order to test whether there is variation across populations in these LLMs. A variation across populations, while controlling for language and register, would indicate that the models are skewed towards specific populations. The first part of the analysis tests whether there are significant differences at a country-level in either (i) the mean perplexity score or (ii) the standard deviation in perplexity scores. This analysis helps us answer the question whether the social media language production of some countries is better represented than others. The second part of the analysis looks at local populations within countries in order to determine if the country-level patterns are robust and then to investigate country-internal variation. Here we ask the question whether some parts of a population are better represented, such as rural vs urban. The third part of the analysis explores whether specific social factors like per capita GDP or amount of international travel are able to predict how well a local population is represented by the models. 

The main contribution of this paper is to measure the influence that geographic population has on the performance of LLMs. The experiments show that neither family of models performs consistently across country-level populations, with North America and the UK being consistently better represented. While there is variation within populations, the results remain stable even at the local level. Further, there is little change across parameter sizes within a single family of models and only a slight difference between families. These experiments show that there is a consistent skew in the performance of LLMs across different populations, what we call a \textit{population skew}. This skew challenges the equity of widely applying LLMs across diverse populations. The supplementary material contains supporting analyses.\footnote{\href{https://doi.org/10.17605/OSF.IO/BZ6PQ}{DOI: 10.17605/OSF.IO/BZ6PQ}}

\section{Related Work}

A tremendous amount of research has focused on probing or evaluating pre-trained LLMs \citep{rogers-etal-2020-primer}. The main focus here is on the degree to which LLMs equally represent different geographic populations using the same language. Given the larger goal of equity in language technology, this is similar to work on multi-lingual probing in that speakers of different languages also constitute different populations. For instance, multi-lingual models tend to work better for high-resource than for low-resource languages \citep{wu-dredze-2020-languages}, zero-shot transfer approaches work much better when the source and target languages have shared vocabulary \citep{deshpande-etal-2022-bert}, and multi-lingual models retain certain preferences from well-represented languages like English \citep{papadimitriou-etal-2023-multilingual-bert}. Such work leaves unasked whether these models perform equally across diverse populations who are using a single language.

\begin{table*}[t]
\centering
\begin{tabular}{|l|l|l|l|l|l|l|l|}
\hline
know & life & via & money & both & amazing & place & least \\
time & thank & yes & free & wait & yet & talking & problem \\
people & well & down & show & women & making & friend & stay \\
day & way & hope & since & end & school & care & covid \\
love & always & god & home & believe & under & power & project \\
new & year & stop & lot & used & anything & once & head \\
see & over & give & nothing & top & coming & wrong & kind \\
think & world & ever & bad & around & state & city & white \\
why & most & feel & find & night & post & working & group \\
here & take & everyone & already & looks & away & nice & health \\
want & man & big & read & family & guy & ready & until \\
go & say & team & through & name & change & times & food \\
really & last & help & part & country & try & business & story \\
\hline
  \end{tabular}
  \caption{Selection of keywords used to select tweets for sub-corpora. Each sub-corpus contains one tweet for each keyword. The supplementary material contains all 250 keywords.}
  \label{tab:keywords}
\end{table*}

Recent work has examined geographic fine-tuning for tasks like dialect modelling which interact with geography \citep{hofmann2023geographic}; other work has shown that multi-lingual fine-tuning can benefit from a geographic selection of languages \citep{nasir-mchechesi-2022-geographical}. In terms of certain populations being under-represented in LLMs, population-specific fine-tuning has been shown to improve the performance of homophobia/transphobia detection \citep{wong2023cantnlpltedi2023} and the performance of hate speech classification more generally has been shown to be inconsistent across different geographic populations \citep{lwowski-etal-2022-measuring}. This is because non-Western biases and disparities are implicitly ignored \citep{ghosh-etal-2021-detecting}. Such findings are perhaps related to the under-representation of scholars from many geographic areas within NLP as a field \citep{rungta-etal-2022-geographic}.

Geographic corpora, like the one used in this paper, have become increasingly common. For instance, the \textit{Corpus of Global Web-based English} \citep{Davies2013} includes web-crawled data, as does the 427 billion word multi-lingual \textit{Corpus of Global Language Use} \citep{Dunn2020} and \textit{GeoWAC} \cite{dunn-adams-2020-geographically}, a collection of gigaword corpora for 50 languages with each language balanced geographically. Geographic corpora explicitly represent different populations of speakers and thus have often been used to study linguistic variation. Such studies also provide a method of validating the corpora themselves: both geographic web corpora \citep{Cook2017} and geo-referenced tweets  \citep{10.3389/frai.2019.00011} can be used to replicate traditional studies of dialectal variation. 

Such work provides a validation of geographic corpora, at least in major languages like English, in the sense that the corpora contain the linguistic variants expected from dialect surveys. Another approach to validating geographic corpora relies on corpus similarity methods \citep{Kilgarriff2001, Li2022a, Li2022} to compare geo-referenced corpora. For instance, recent work has shown that there is a consistency across languages between geo-referenced corpora from the web and from social media \citep{Dunn2021}. In spite of this general validity, there are still confounds present in geo-referenced tweets \citep{pavalanathan-eisenstein-2015-confounds}, although these confounds do not change the underlying idea in this paper that different places represent different populations. The presence of confounds does mean, however, that geographic corpora do not always represent the full range of social variation within each local population.

\begin{figure*}[t]
\centering
\includegraphics[width =.98\textwidth]{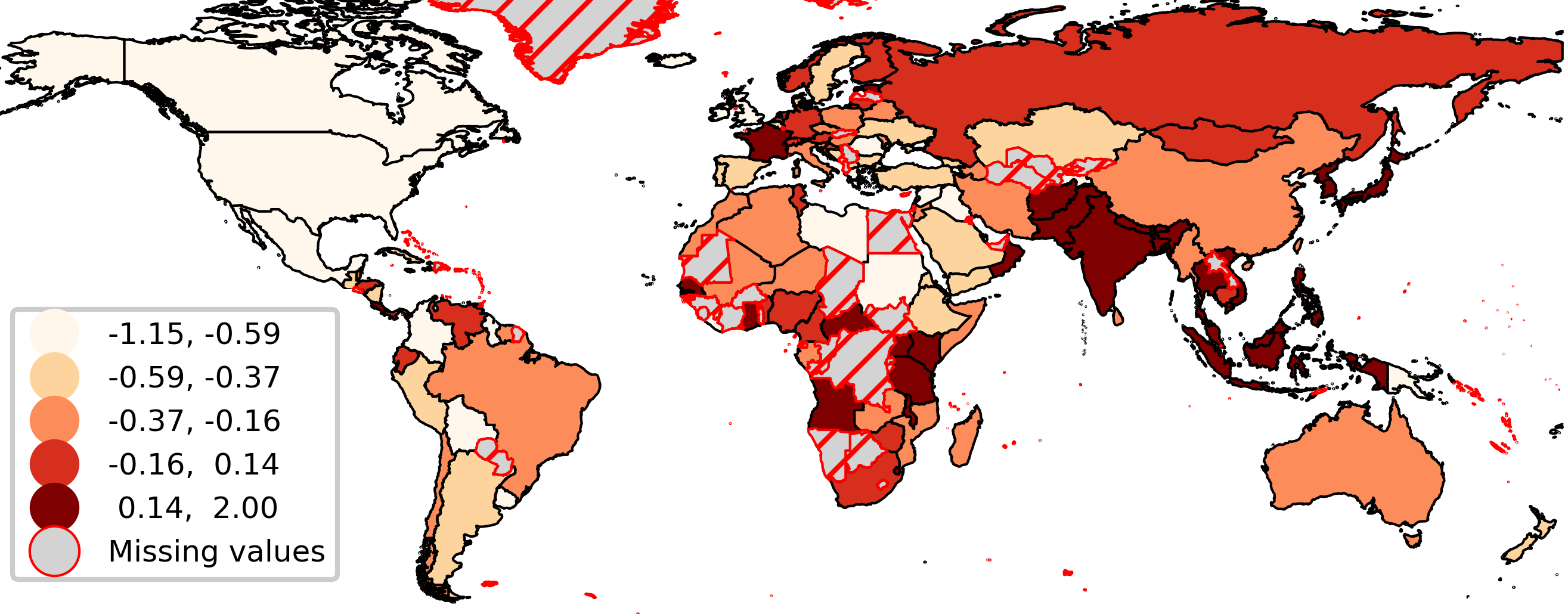}
\caption{Countries by standardized perplexity scores across sub-corpora for BigScience \textsc{bloom} 3b. Because the z-score is used, 1 reflects one standard deviation above the mean and -1 below the mean.}
\label{fig:map_perplexity}
\end{figure*}

\section{Language Data}

The data set used for these experiments is drawn from geo-referenced tweets. The underlying collection is drawn from a set of 927 airports around the world, with each airport representing the surrounding metropolitan area (within a 25km radius). While data is collected from a larger set of 15k cities\footnote{\href{https://www.geonames.org/}{https://www.geonames.org/}}, only these 927 have enough data to form at least one sub-corpus once the selection criteria are fulfilled. We select English language samples, as identified by two language identification systems \citep{Dunn2020, dunn-nijhof-2022-language}. The distribution of these local populations is shown in Figure \ref{fig:map}. This data is spread widely enough to undertake a study of different populations around the world but not to provide an exhaustive representation.


\begin{table}[h]
\centering
\begin{tabular}{|l|rr|}
\hline
\textbf{Region} & \textbf{Places} & \textbf{Sub-Corpora} \\
\hline
Africa, North & 33 & 929 \\
Africa, South & 20 & 3,366 \\
Africa, Sub & 67 & 5,036 \\
\hline
America, Brazil & 13 & 57 \\
America, South & 92 & 1,237 \\
America, Central & 63 & 4,403 \\
America, North & 96 & 11,331  \\
\hline
Asia, Central & 38 & 1,171 \\
Asia, East & 62 & 2,422 \\
Asia, South & 81 & 13,889 \\
Asia, Southeast & 83 & 4,500 \\
\hline
Europe, East & 70 & 5,274 \\
Europe, West & 173 & 20,100 \\
Europe, Russia & 32 & 112 \\
\hline
Middle East & 38 & 1,929 \\
Oceania & 36 & 10,430 \\
\hline
\textbf{Total / Avg} & \textbf{974} & \textbf{86,186} \\
\hline
  \end{tabular}
  \caption{Distribution of Sub-Corpora by Region. Each sample is a unique sub-corpus with the same distribution of keywords, with one tweet for each keyword, for an average of 3,910 words.}
  \label{tab:1}
\end{table}

The main challenge is to control for other sources of variation like topic or register that would lead to variation in the results of the probing task (i.e., higher perplexity) but would not be directly connected with the local population being observed. In other words, we need to constrain the production of the local populations to a specific set of topics. For this reason, we develop a set of 250 common lexical items which are neither purely functional (like \textit{the} is) nor purely topical (like \textit{Biden} is). A selection of these keywords is shown in Table \ref{tab:keywords} and the complete list is provided in the supplementary material. For each location we create sub-corpora which are composed of one unique tweet for each of these keywords. Thus, each location is represented by a number of sub-corpora which each have the same fixed distribution of key lexical items. This controls for wide variations in topic or content or register, factors that would otherwise create unnecessary lexical variation. Thus, we expect each sub-corpus to have approximately the same perplexity for a given LLM.

This creates a data set with 86,186 sub-corpora, each with an average of approximately 3,910 words. The distribution of sub-corpora is show in Table \ref{tab:1}, with locations divided into regions. For example, in South America there are 1,237 samples (sub-corpora) distributed across 92 local metropolitan areas. While wealthy western countries in North America and Western Europe provide the single largest numbers of local populations, 72\% of local areas come from outside these two regions as well as 64\% of sub-corpora. Thus, this data set provides corpora which are comparable in terms of topic and register and represent a diverse set of English-using populations around the world. While English is not the primary language in all locations, it is remains widely used in each.

\section{Methods}

\begin{figure*}[t]
\centering
\includegraphics[width = .95\textwidth]{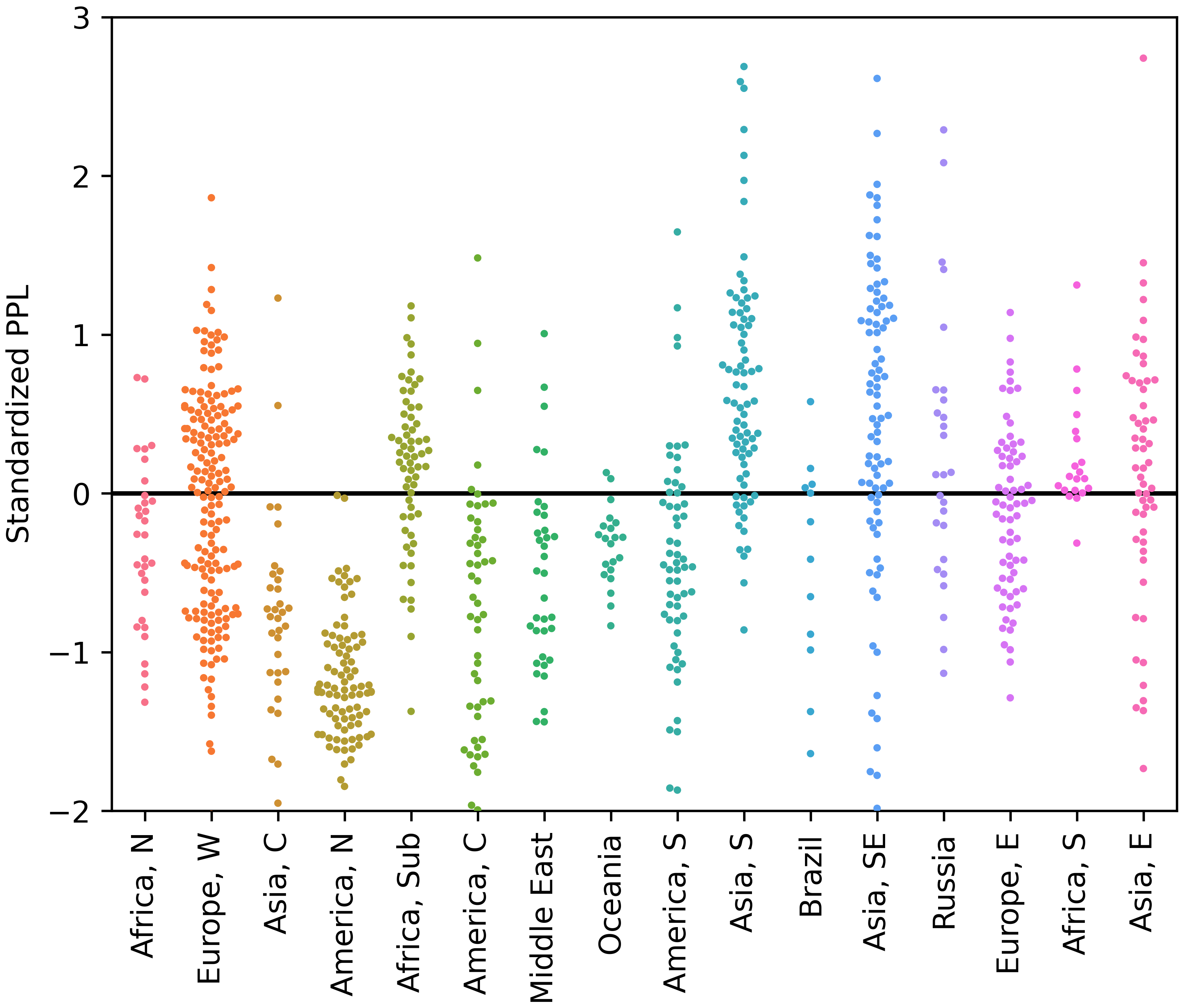}
\caption{Local populations grouped by region by standardized perplexity scores for Facebook \textsc{opt} 2.7b}
\label{fig:swarm}
\end{figure*}

For this probing task, the question is whether these LLMs consider the sub-corpora from each population to be equally likely. This is measured using perplexity, capturing the goodness-of-fit for each model for each sub-corpus. For example, if the LLM has a bias towards speakers of American English then we would expect a lower perplexity score for samples from the US and Canada and a higher perplexity score for samples from India and Pakistan. This goodness-of-fit task is a direct view of the representativeness of the models for each local population. However, it is imprecise from a linguistic perspective because the perplexity scores could reflect differences in linguistic knowledge (e.g., syntax and lexis) or in world knowledge (e.g., local entities and place names). All reported perplexity scores are standardized using the z-score across all samples for each model; thus, 0 is the global mean and a value of -1 would be one standard deviation below the mean for that model. We are not concerned with the absolute goodness-of-fit for each model, but rather in the distribution of such measures across different geographic populations.

We use three models each from two families of LLMs to carry out this task: the BigScience \textsc{bloom} series (with 560m, 1.7b, and 3b parameters: \citealt{workshop2023bloom}) and the Facebook \textsc{opt} series (with 350m, 1.7b, and 3b parameters: \citealt{zhang2022opt}). This provides a view of whether one family has more or less population skew as well as whether larger parameter sizes have more or less population skew. Both the \textsc{bloom} and \textsc{opt} series are based on decoder-only transformer architectures using a causal language modelling task. The hyperparameters are likewise comparable: all models have 24 layers, except \textsc{opt} 2.7b which has 32 and \textsc{bloom} 3b which has 30; the number of heads is the same, at 16 for the smaller and 32 for the larger parameter sizes. Embedding sizes range from 1024 (in the 560m and 350m models) to 2560 (in the 3b and 2.7b models). 

Thus, we would expect that any differences in these models result from differences in the training corpora because the architectures are largely equivalent. The \textsc{opt} series is trained on English data alone while \textsc{bloom} includes 46 natural and 13 programming languages. Both series favor digital corpora, a result of the need for sufficient data. While we do not have information about the geographic distribution of the training data for either family, the probing task undertaken here provides an indication of where the training data is drawn from insofar as populations which are better represented upstream are likely to be better represented downstream.

\section{Results and Analysis}

We begin the analysis by mapping the mean perplexity score by country, shown in Figure \ref{fig:map_perplexity}. Here darker red means a higher perplexity which in turn means a lower fit between the model and the sub-corpora from that country. Thus, we see that there is a strong geographic skew towards North America and the UK, with those countries in the group which has the lowest perplexities. This includes Mexico as well, a country in which English is not the majority language. Other countries that are traditionally viewed as inner-circle speakers of English \citep{k90} have higher perplexity scores: New Zealand is in the next category up and Australia is two categories up. In fact, countries like Argentina, Spain, and Mongolia are better represented in the model than Australia.

\begin{figure}[b]
\centering
\includegraphics[width=\columnwidth]{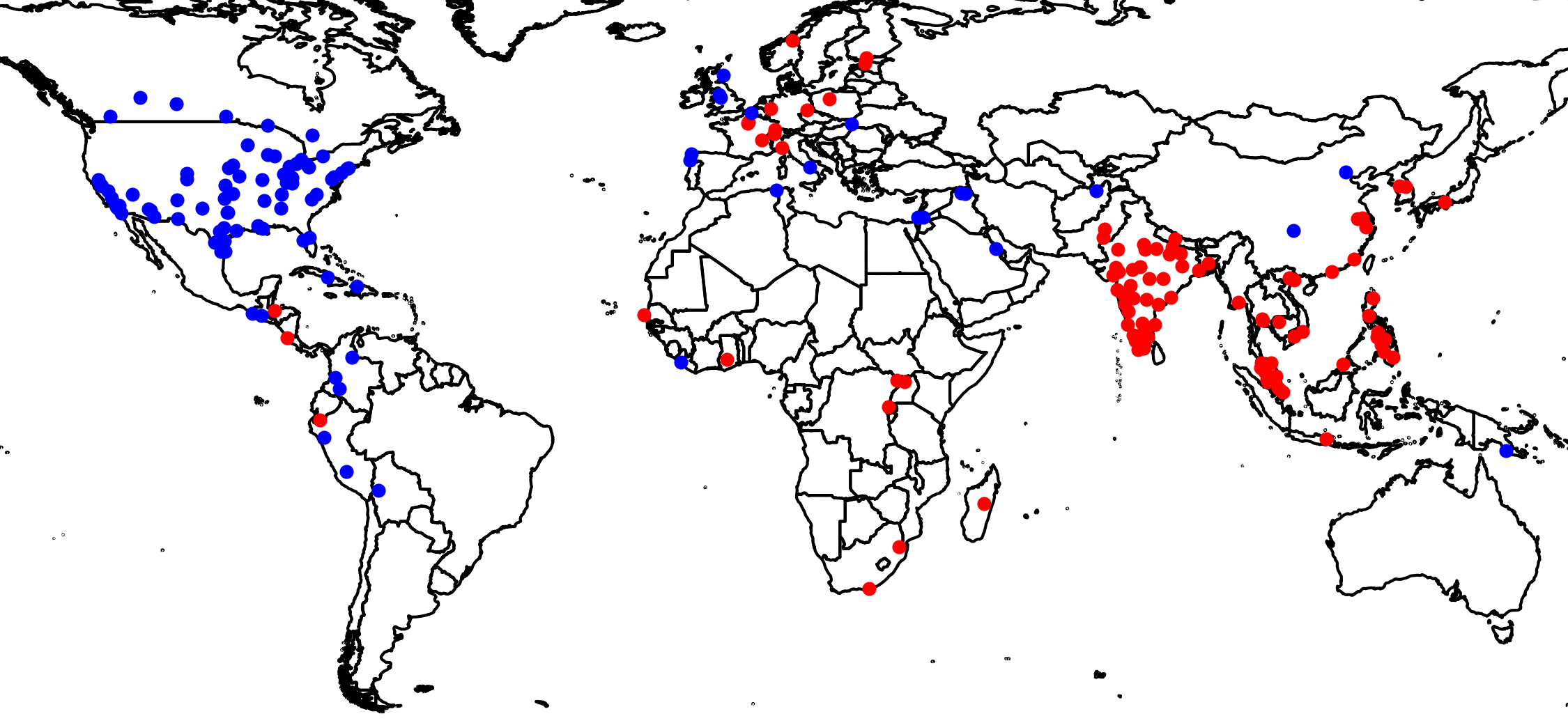}
\caption{Top-100 lowest (blue) and highest (red) mean perplexity locations, Facebook \textsc{opt} 2.7b.}
\label{fig:highlowfb}
\includegraphics[width=\columnwidth]{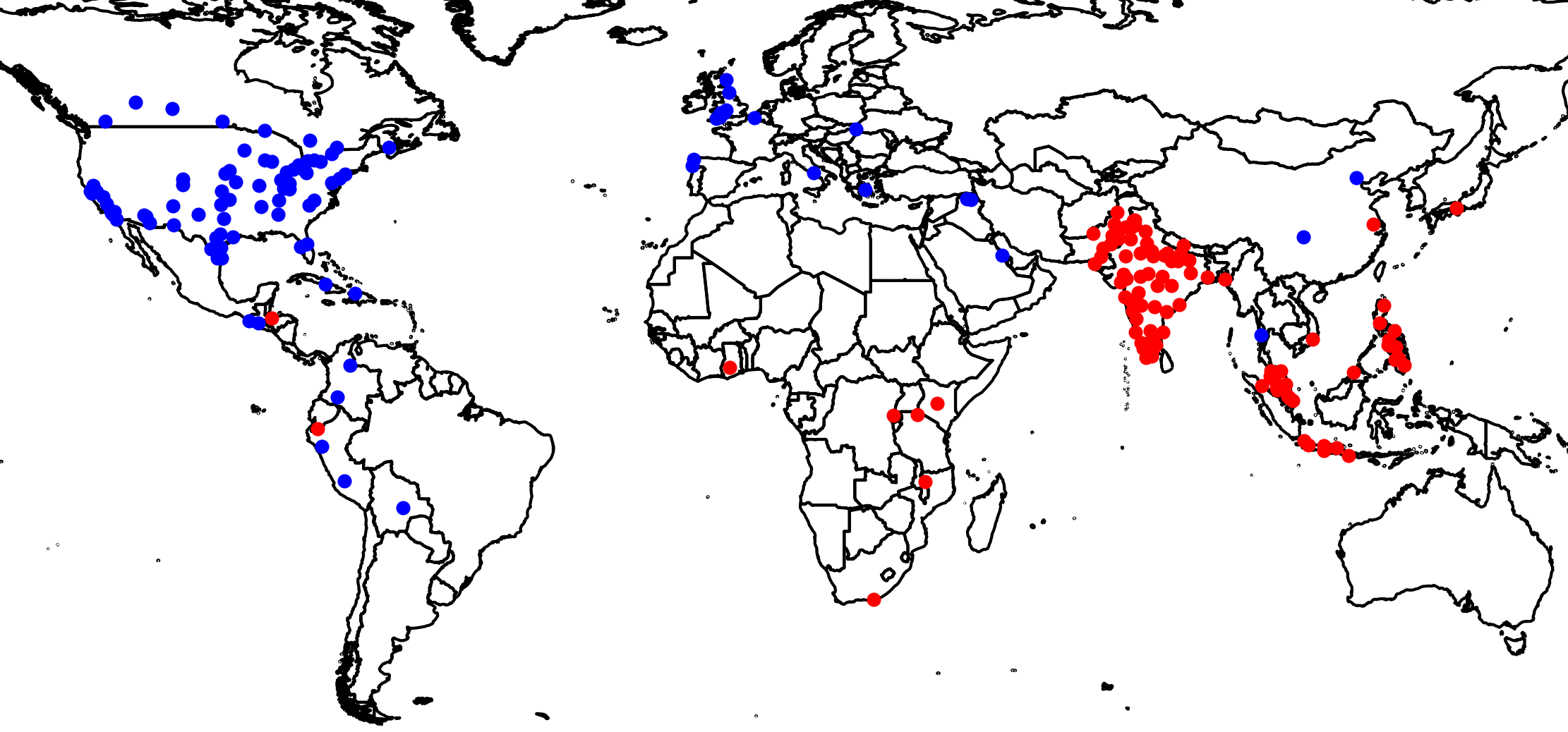}
\caption{Top-100 lowest (blue) and highest (red) mean perplexity locations, BigScience \textsc{bloom} 3b.}
\label{fig:highlowbloom}
\end{figure}

Figures~\ref{fig:highlowfb} and \ref{fig:highlowbloom} show a different perspective by mapping cities where the average perplexity is either the lowest or the highest. We first filter cities to remove those with fewer than 10 sub-corpora. Blue locations are the top-100 lowest mean perplexity cities and red locations are the top-100 highest mean perplexity cities. The results show that the majority of lowest-perplexity cities are in North America, with the very lowest perplexity cities including a number of locations in the middle and southwest United States. In contrast, for highest perplexity, the majority are in India and Southeast Asia. Perhaps most remarkable is that, apart from some high-perplexity cities in Europe for \textsc{opt}, there is a great deal of overlap in the highest and lowest perplexity locations for the \textsc{opt} and \textsc{bloom} models.

Local populations within countries are not all equally represented, however. The swarm plot in Figure \ref{fig:swarm} shows a different view, with each local population a point that is plotted according to its standardized perplexity. Thus, 0 here (marked by a black line) is the average perplexity across the entire data set. Some regions have most of their local populations below the line, thus indicating a consistently better fit for the model (e.g., North America has consistently good fit). Other areas, like South Asia, have most of their local populations above the average, showing a consistently poor fit. A number of regions, though, show internal variation; for instance, Western Europe straddles the line with many of its local areas having a good fit and many others having a poor fit. Thus, there is not a strict consistency at the regional level.

\begin{figure*}[t]
\centering
\includegraphics[width = .98\textwidth]{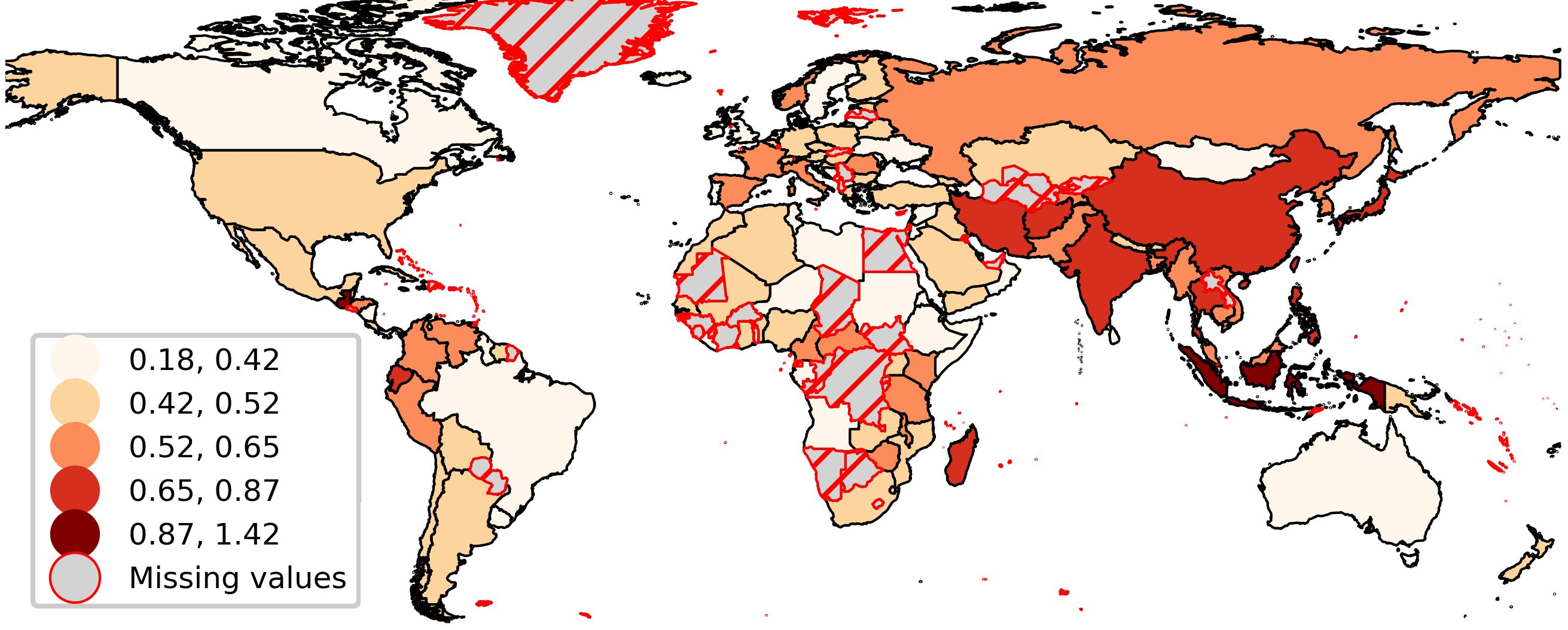}
\caption{Countries by standard deviation in standardized perplexity scores across sub-corpora for BigScience \textsc{bloom} 3b. Higher standard deviations indicate more variation within a country.}
\label{fig:map_std}
\end{figure*}

\begin{table}[h]
\centering
\begin{tabular}{|l|cc|}
\hline
\textbf{Country} & \textbf{Std. Dev.} & \textbf{N. Locations} \\
\hline
United Kingdom & 0.19 & 26 \\
Australia & 0.22 & 15 \\
Nigeria & 0.27 & 11 \\
Canada & 0.31 & 24 \\
Germany & 0.34 & 19 \\
\hline
\hline
Thailand & 1.09 & 13 \\
Colombia & 1.10 & 17 \\
Ireland & 1.16 & 13 \\
India & 1.29 & 54 \\
Indonesia & 1.96 & 25 \\
\hline
  \end{tabular}
  \caption{Standard deviation of average perplexity (standardized) by country; more homogeneous countries are on the top and more heterogeneous countries on the bottom.}
  \label{tab:stddev}
\end{table}

Another perspective is shown in Table \ref{tab:stddev} with the most homogeneous countries across local areas (top) and the most heterogeneous (bottom). The measure here is the standard deviation of average perplexity scores. More developed western countries tend to be more homogeneous; for example the table shows the United Kingdom with a standard deviation of 0.19 and Australia of 0.22. On the other hand, countries like India and Indonesia have much more variation internally (1.29 and 1.96, respectively). There is not a complete demographic separation, however, as Ireland also shows a relatively high standard deviation. The point here is that, while in the aggregate some countries are better represented than others, there is also variation within many countries.

A wider view of country-internal variation is shown in Figure \ref{fig:map_std}; this is a comparable map to Figure \ref{fig:map_perplexity} except that it shows the range of variation within a country rather than the mean perplexity. Thus, here darker red countries are subject to more variation in how well the model represents the local population. Here Canada, Brazil, and Australia have less internal variation (regardless of how well the model fits the population). On the other hand, the largest internal variation is seen in China, India, and the Middle East. This indicates that there are portions of the population of these countries which are poorly represented and others that are better represented, a matter of country-internal diversity.

\begin{figure}[b]
\centering
\includegraphics[width = \columnwidth]{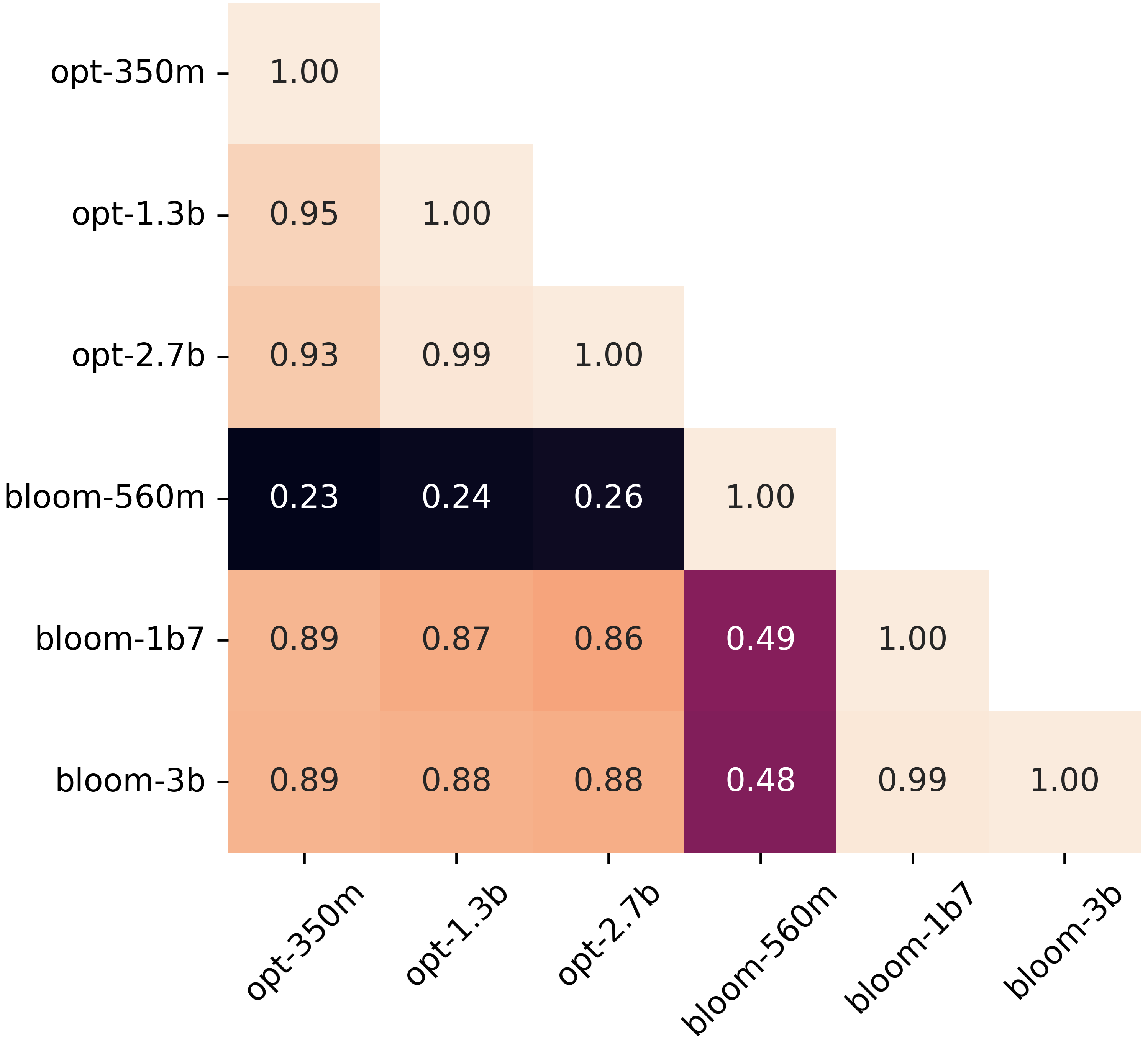}
\caption{Pearson correlations between standardized perplexity scores across models}
\label{fig:heatmap}
\end{figure}

To what degree does the model family or size change this population skew? This is shown using a heatmap in Figure \ref{fig:heatmap}. Within each family, there is high agreement. Within \textsc{opt} there are correlations of 0.93 to 0.99 between parameter sizes, while only the \textsc{bloom} 560m model is quite different from the other \textsc{bloom} models. Across families (i.e., \textsc{opt} vs \textsc{bloom}) agreement is still quite high, although now in the upper .80s. Only the smallest model in the \textsc{bloom} family is an outlier here, and it is an outlier in its relations to all other models. This figure shows, then, that the overall pattern of population skew is remarkably consistent across both model family and parameter size. This is important because it shows that the size of the model and the specific set of training data (a factor on which the two families differ) do not change the overall skew found in these LLMs.

\begin{table}[h]
\centering
\begin{tabular}{|llc|}
\hline
\multicolumn{3}{|c|}{\textbf{OPT 2.7b}} \\
\textit{Country} & \textit{City} & \textit{P Value} \\
\hline
China & Hong Kong & 0.02 \\
Pakistan & Karachi & 0.02 \\
Pakistan & Lahore & 0.03 \\
Germany & Düsseldorf & 0.01 \\
United Kingdom & South London & 0.03 \\
\hline
\hline
\multicolumn{3}{|c|}{\textbf{BLOOM 3b}} \\
\textit{Country} & \textit{City} & \textit{P Value} \\
\hline
Afghanistan & Kabul & 0.02 \\
Pakistan & Lahore & 0.04 \\
Germany & Düsseldorf & 0.03 \\
Georgia & Tbilisi & 0.04 \\
\hline
  \end{tabular}
  \caption{Local areas with a significant difference in samples by time. 97.4\% of local areas are consistent across samples using a two-tailed t-test.}
  \label{tab:robustness}
\end{table}

\begin{figure}[b]
\centering
\includegraphics[width = \columnwidth]{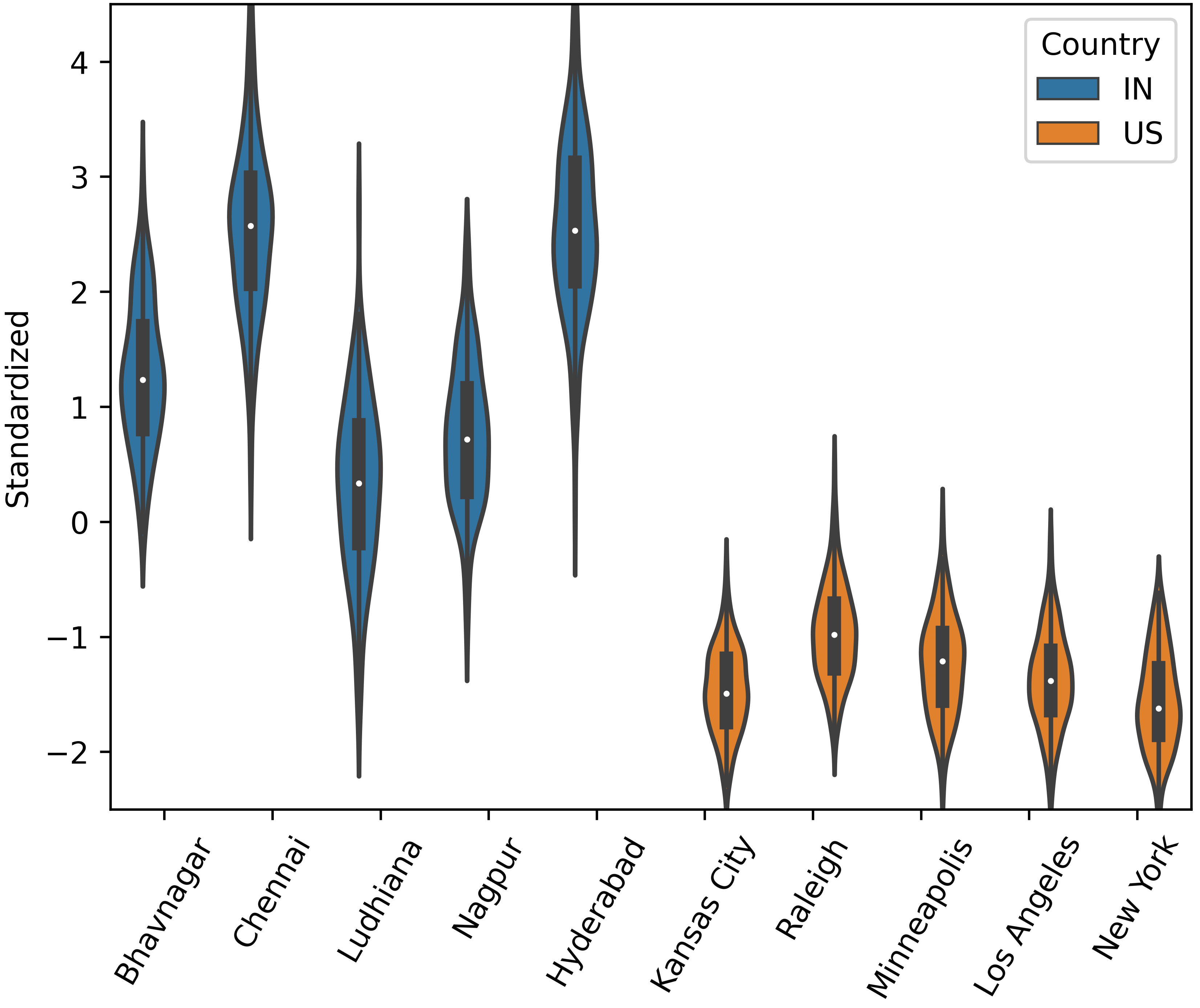}
\caption{Distribution of perplexity values for select cities in the US and India, Facebook \textsc{opt} 2.7b}
\label{fig:cities}
\end{figure}

We take a closer look at local differences in Figure \ref{fig:cities} with five cities each from India and the US. These violin plots show the whole distribution of perplexity scores across samples, thus showing both the level of fit and the level of internal variation at the same time. First, it is clear that the model (\textsc{opt 2.7b}) favors the US. Second, it is also clear that samples from the US are more consistent or homogeneous within each city. There is some variation within countries, for example with the more southern city of Raleigh having a higher perplexity than other US cities; but this difference is minor compared to the country-level differences.

To what degree is this analysis robust to the specific samples being observed? We evaluate this using the set of local populations which are relatively well-represented by having more than 150 distinct sub-corpora. For these local areas we divide the sub-corpora into two groups by time period and then use a two-tailed t-test to determine if there is any difference in these two independent populations of sub-corpora. Essentially, the question here is whether the results would differ given a different set of samples from each local population. Out of 153 local areas with sufficient samples to conduct this test, only five show a significant difference for \textsc{opt} 2.7b and four for \textsc{bloom} 3b. These are shown in Table \ref{tab:robustness}. Most of these locations either have under-represented local populations or are in border areas. Still, 97.4\% of local areas show no difference in the distribution of perplexity values when evaluated on unique sets of samples. This gives us confidence that the population skew shown above is not an arbitrary finding that depends on the specific sub-corpora being observed.

\begin{table}[!h]
\centering
\begin{tabular}{|cccc|}
\hline
\textbf{Circle} & \textbf{Country} & \textbf{\textsc{opt 2.7b}} & \textbf{\textsc{bloom 3b}} \\
\hline
Inner & AU & -0.23 & -0.34 \\
Inner & CA & -0.79 & -0.86 \\
Inner & IE & -0.61 & -0.73 \\
Inner & NZ & -0.20 & -0.46 \\
Inner & UK & -0.84 & -0.89 \\
Inner & US & -1.17 & -1.01 \\
\hline
Between & ZA & 0.12 & -0.01 \\
\hline
Outer & BD & 0.39 & 0.45 \\
Outer & IN & 1.06 & 1.13 \\
Outer & KE & 0.21 & 0.56 \\
Outer & MY & 1.38 & 1.16 \\
Outer & NG & 0.22 & 0.06 \\
Outer & PH & 1.01 & 1.52 \\
Outer & PK & 0.11 & 0.90 \\
Outer & TZ & 0.07 & 0.15 \\
\hline
Expanding & CN & -0.04 & -0.29 \\
Expanding & FR & 0.61 & 0.15 \\
Expanding & JP & 0.20 & 0.54 \\
Expanding & KR & 0.16 & 0.19 \\
Expanding & NL & 0.44 & -0.06 \\
Expanding & NP & 0.15 & 0.27 \\
Expanding & RU & 0.10 & -0.04 \\
\hline
  \end{tabular}
  \caption{Mean standardized perplexity by country, organized by circles of English.}
  \label{tab:circles}
\end{table}

One approach to differentiating between the status of different populations of English speakers by country is Kachru's \textit{circles} model, based on the historical pattern of colonization \cite{k90}. \textit{Inner-circle} countries like the US or Australia are the first diaspora, where English was transplanted by settler-speakers. \textit{Outer-circle} countries like Kenya or India are the second diaspora, where English was transplanted by imperial expansion. And \textit{Expanding-circle} countries are the third diaspora, where English is learned as a \textit{lingua franca}, for instance for business purposes, rather than through direct colonization. The mean perplexity by country organized in this fashion is shown in Table \ref{tab:circles}. In general, inner-circle countries have the best fit (i.e., the lowest perplexity). However, many expanding-circle countries are lower than outer-circle countries. Thus, there are more factors involved here than simply the historical status of English in each of these countries. For instance, the Netherlands is a wealthy and well-connected Western country with a relatively low perplexity in both models, lower than many outer-circle countries. Thus, historical factors cannot fully explain the distribution of perplexity values.

\begin{figure*}[ht]
\centering
\includegraphics[width = 450pt]{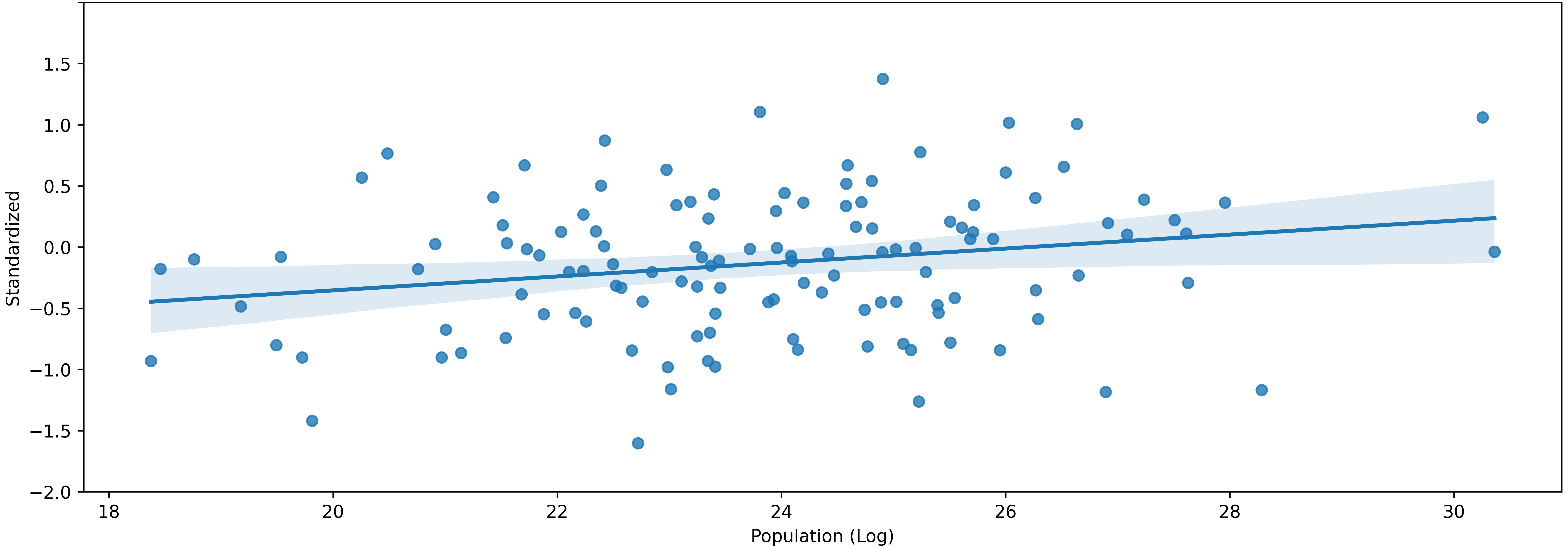}
\caption{Regression plot showing the relationship between mean perplexity (y-axis) and size of population (x-axis) for local populations; Facebook \textsc{opt} 2.7b}
\label{fig:reg_pop}

\includegraphics[width = 450pt]{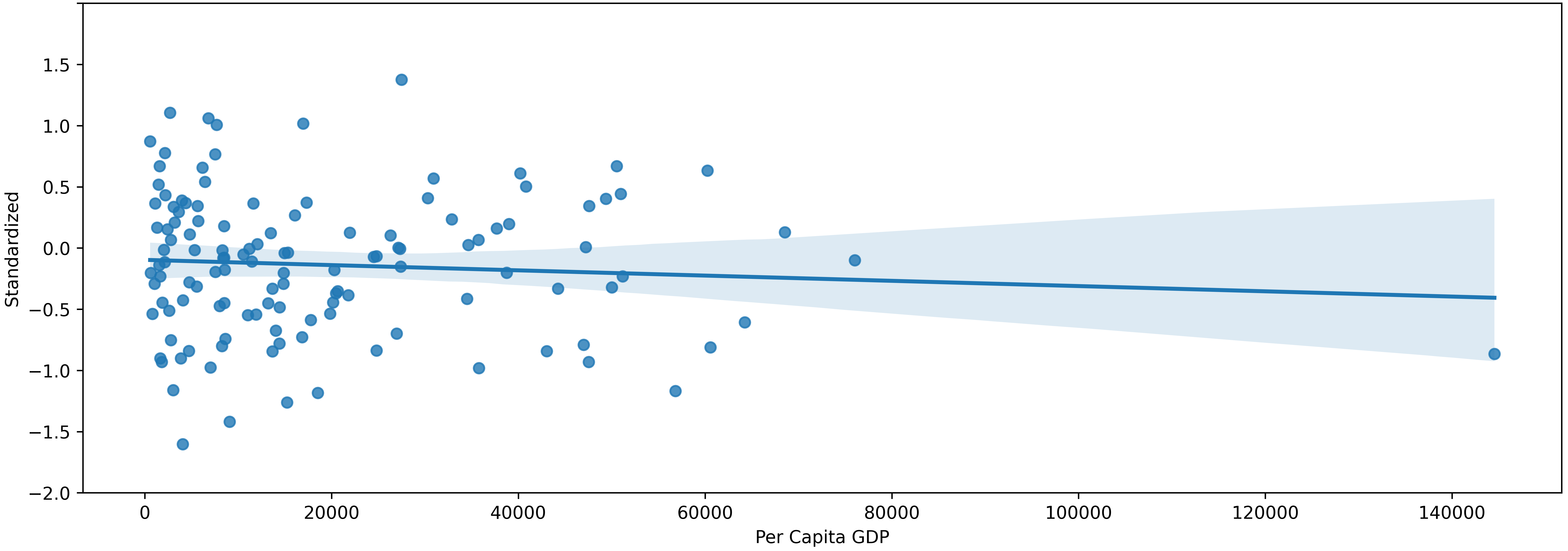}
\caption{Regression plot showing the relationship between mean perplexity (y-axis) and per capita GDP (x-axis) for local populations; Facebook \textsc{opt} 2.7b}
\label{fig:reg_gdp}
\end{figure*}

The next question is whether there is a predictable relationship between the size of a country's population or its relative wealth and the degree to which the LLMs well-represent that country on average. This is shown in two regression plots in Figures \ref{fig:reg_pop} and \ref{fig:reg_gdp}. In the first, we see no clear relationship between population size and fit with the model. There is a slight upward trend (where higher perplexity means a worse fit), but this is largely an artifact of the extreme size of China and India, both with relatively poor fits. The log population is shown in order to make the figure more readable. Thus, at the country level, it is not the case that simply having more people improves a country's representation.

Another possibility is that wealthier countries, with greater degrees of internet access and thus higher contribution to digital corpora, would have a better fit with the model. This is shown in Figure \ref{fig:reg_gdp}, using per capita GDP. Again there is no clear relationship between the two; there is a slight downward trend caused by a single outlier. Thus, it is not the case that wealth alone drives the degree of fit across these population-specific sub-corpora.

These two regression plots aggregate by country, even though we know there is some variation within countries. We therefore undertake an analysis of all 974 local populations as well. Here we use the amount of international air travel as a proxy for the connectedness of each local population, testing the hypothesis that those locations which are better connected will have more influence on the model. The amount of air travel is taken from previous work which provides estimates at the level of airports \citep{10.1371/journal.pone.0064317}. Because this analysis includes all local populations, there are many more observations. There is again no real relationship here, so that more internationally-connected populations (representing larger urban centers) are not better described by these models.

These findings are troubling because they show (i) that pre-trained language models have a strong geographic skew, (ii) that this skew is robust and persistent, not an artifact of a specific corpus, but (iii) that while some social factors are related to the skew, the skew itself remains unpredictable. Systematic sources of error can be systematically corrected. But such arbitrary fluctuations by population call into question the idea that these language models are adequate for representing language use even within a single language and register.

\section{Discussion and Conclusions}

In this paper we presented a spatial probing task on the BigScience \textsc{bloom} and Facebook \textsc{opt} models to measure differences in the perplexities of samples of social media data collected from 927 local populations. Our results have shown conclusively that these two families of LLMs, with similar architectures but trained on different corpora, share a pervasive geographic population skew. The results show that some populations are much better represented than others. This is the case even when the probing task is constrained to a single language and register in order to eliminate confounds related to multi-lingual representations. While this skew itself is quite clear, it is not predictable given social or economic factors. This means that we do not yet know why these models perform better on some populations than others.

One potential cause is the geographic distribution of the training corpora. However, current models lack an explicit characterization of the geographic sources of training data. Thus, we can only indirectly capture patterns in the training data using probing tasks such as the one described in this paper. In the long term, language mapping projects like earthLings.io\footnote{\href{https://www.earthLings.io}{https://www.earthLings.io}} offer a way to more fully understand the sources of training data. It should be noted, though, that the geographic skew shown here is highly correlated across model families even though these families rely on different sets of training data. This tells us two important facts: (i) that these models are unable to generalize from language use representing specific populations to language as a whole and (ii) that the geographic distribution of training data is an unknown quantity in these models.

Given the current trend toward using a single pre-trained LLM as the starting point for language technologies, this means that many populations will be inequitably treated unless some explicit population adaptation is undertaken.  Although the scale of many LLMs make it difficult to understand downstream effects, we have shown that perplexity provides a relatively lightweight way to understand skew in the representation of geographic populations. This type of analysis can be applied to other models, including closed source ones. Geographic skew acts as an indicator for possible bias in LLMs because demographic characteristics tend to be spatially clustered, and therefore geography provides a lens into the multidimensional nature of human populations \cite{cliff1970spatial}.

\section{Ethics Statement}

This paper uses written digital corpora to represent diverse populations of speakers around the world. While such geographic corpora are more representative of the world's population than non-geographic corpora, it remains the case the certain demographic segments within each geographic area are more likely to contribute to the corpus. Thus, these results should not be taken as an exhaustive representation of all sub-populations within each geographic location.

\section{Bibliographical References}\label{sec:reference}

\bibliographystyle{lrec-coling2024-natbib}
\bibliography{lrec-coling2024-example}

\end{document}